\documentclass{IEEEcsmag-arxiv}
\usepackage[colorlinks,urlcolor=blue,linkcolor=blue,citecolor=blue]{hyperref}
\expandafter\def\expandafter\UrlBreaks\expandafter{\UrlBreaks\do\/\do\*\do\-\do\~\do\'\do\"\do\-}
\usepackage{upmath,color}

\jvol{XX}
\jnum{XX}
\paper{x}
\jmonth{May}
\jname{}
\jtitle{Ethical ChatGPT: Concerns, Challenges, and Commandments}
\pubyear{2023}

\begin{document}

\sptitle{}

\title{Ethical ChatGPT: Concerns, Challenges, and Commandments}

\author{Jianlong Zhou}
\affil{Data Science Institute, University of Technology Sydney, Sydney, NSW 2007, Australia}

\author{Heimo Müller}
\affil{Human-Centered AI Lab, Medical University Graz, Austria}

\author{Andreas Holzinger}
\affil{Human-Centered AI Lab, University of Natural Resources and Life Sciences Vienna and Medical University Graz, Austria} 

\author{Fang Chen}
\affil{Data Science Institute, University of Technology Sydney, Sydney, NSW 2007, Australia}

\markboth{THEME/FEATURE/DEPARTMENT}{THEME/FEATURE/DEPARTMENT}

\begin{abstract}
Large language models, e.g. ChatGPT are currently contributing enormously to make artificial intelligence even more popular, especially among the general population. However, such chatbot models were developed as tools to support natural language communication between humans. Problematically, it is very much a ``statistical correlation machine" (correlation instead of causality) and there are indeed ethical concerns associated with the use of AI language models such as ChatGPT, such as Bias, Privacy, and Abuse. 
This paper highlights specific ethical concerns on ChatGPT and articulates key challenges when ChatGPT is used in various applications. Practical commandments for different stakeholders of ChatGPT are also proposed that can serve as checklist guidelines for those applying ChatGPT in their applications. These commandment examples are expected to motivate the ethical use of ChatGPT.


\end{abstract}

\maketitle

\chapteri{C}hat Generative Pre-Trained Transformer (also known as ChatGPT), can fluently answer questions from users and has the ability to generate human-like text with a seemingly logical connection between different sections. Individuals have reportedly used ChatGPT to formulate university essays, scholarly articles with references \cite{liebrenz2023generating}, debug computer program code, compose music, write poetry, give restaurant reviews, generate advertising copy and solve exams \cite{elon_2023}, co-author journal articles \cite{pavlik2023collaborating} and many others.

ChatGPT models are basically massive neural networks with billions of parameters, which resulted in gains in quality, accuracy, and breadth of generated content. Their behaviors are learned from a large amount of text data of Internet resources such as web pages, books, research articles and
social chatter, not programmed explicitly. They are trained with two phases: 1) the initial ``pre-training'' phase learns to predict the next word in a sentence with a large amount  of Internet text from a vast array of perspectives; and 2) the second phase ``fine-tunes'' models with the use of datasets that human reviewers crafted to narrow down system behavior \cite{openai_2023}. Such combination of unsupervised pre-training and supervised fine-tuning helps to generate human-like responses to queries and in particular provide responses to queried topics that resemble that of a human expert \cite{floridi2020gpt}.

The rapid widespread adoption of ChatGPT after release has demonstrated its tremendous powerfulness of potential uses in different areas ranging from technical assistance such as coding, essay writing, business letters, to customer engagement as well as many others \cite{tung_chatgpt_2023}.
Despite the powerful capacity of ChatGPT to help people with various writing tasks and experiments engendering both positive and adverse impacts, the society has critical concerns on allowing users to cheat and plagiarize especially in academy and education communities, potentially spreading misinformation, and enabling unethical business practices as well as other ethical issues \cite{stokel2023chatgpt}.

Weidinger et al. \cite{weidinger2021ethical} summarises the ethical risk landscape with Large Language Models (LLM),  identifying six ethical concerns: 1) Discrimination, Exclusion, and Toxicity, 2) Information Hazards, 3) Misinformation Harms, 4) Malicious Uses, 5) Human-Computer Interaction Harms, and 6) Automation, Access, and Environmental Harms. 
ChatGPT shares not only the similar ethical issues with other AI solutions including fairness, privacy and security, transparency, accountability, etc. \cite{zhou2020survey,zhang_one_2023}, it may also introduce additional ethical concerns because of its specific characteristics. For example, people have difficulty to distinguish facts and fake with the ChatGPT's human-like conversations; in education, teachers may have difficulty to differentiate the authorship between human and AI in home work; in the creative areas such as designing and creative writing, ChatGPT may introduce changes to not only the authorship, but also the creativity of human in the long time.

This paper first highlights specific ethical concerns on ChatGPT and articulates key challenges when ChatGPT is used in various applications. We then propose practical commandments for different stakeholders of ChatGPT that can serve as checklist guidelines for those applying ChatGPT in their applications.

\section{ETHICAL CONCERNS}
\begin{figure*}
\centerline{\includegraphics[width=0.95\linewidth]{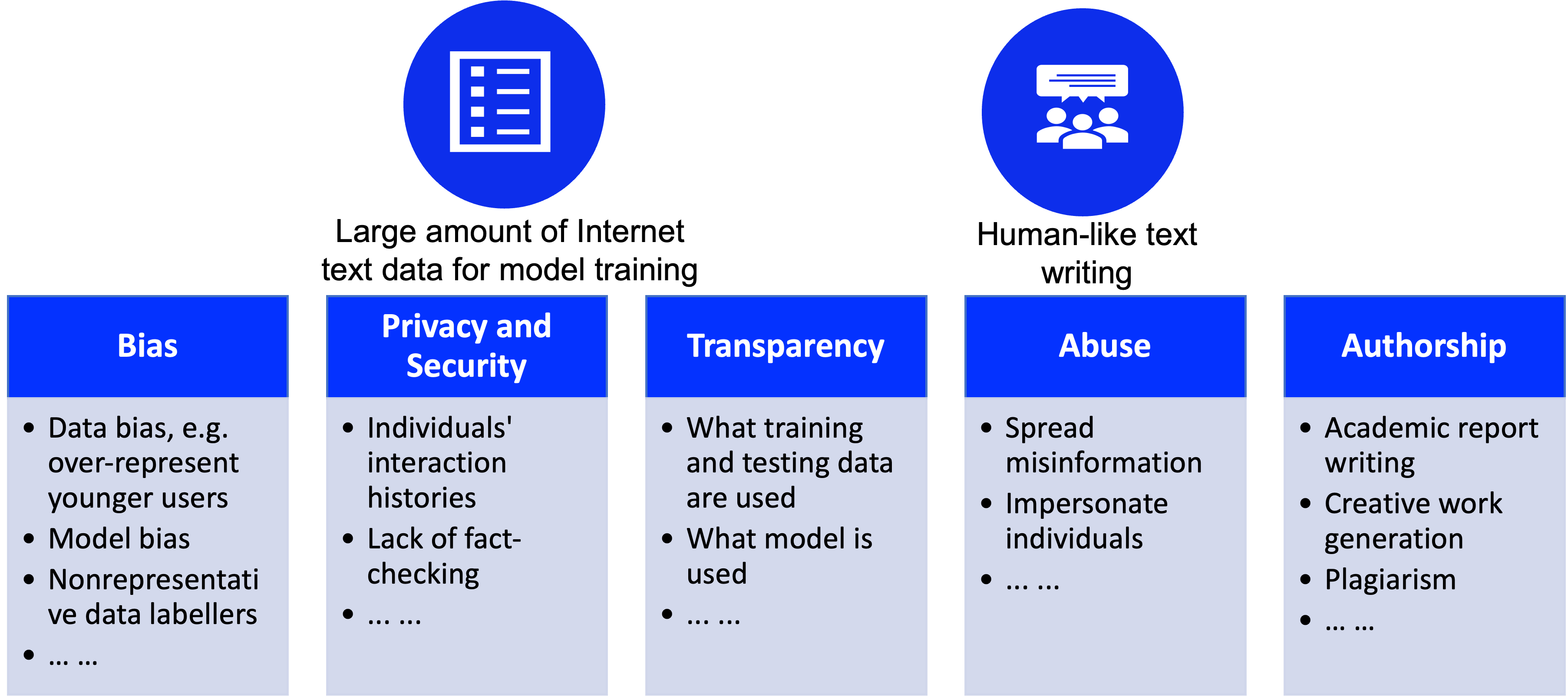}}
\caption{Examples of ethical concerns on ChatGPT}\vspace*{-5pt}
\label{fig:ethical-concerns}
\end{figure*}

This section demonstrates typical ethical concerns on ChatGPT as shown in 
Figure~\ref{fig:ethical-concerns}.

\subsection{Bias}

Similar to many other AI solutions, ChatGPT could also demonstrate bias in its answers. These biases have arisen because of different reasons such as the machine learning algorithms used for modelling and the data used for training and fine-tuning. Despite the use of human data labellers following instructions by ChatGPT for training and fine-tuning datasets, it must be recognised that the data labellers are not representative for diverse viewpoints and perspectives which introduces biases to data. Furthermore, training data is primarily from massive Internet resources which not only have limited diversity but also may have biases within itself. For example, \cite{bender2021dangers} showed that such large datasets significantly over-represent younger users, especially people from developed countries and English speakers.
Any biases present in these data will be reflected in the output of the model. Such bias is hard to overcome \cite{chan2023gpt}. OpenAI lists this issue in its announcement blog post saying that ChatGPT is ``often excessively verbose and overuses certain phrases, such as restating that it’s a language model trained by OpenAI. These issues arise from biases in the training data (trainers prefer longer answers that look more comprehensive) and well-known over-optimization issues''.

These biases can result in unavoidable unfair results of ChatGPT answers, particularly for vulnerable groups.


\subsection{Privacy and Security}
ChatGPT generates answers based on the input it receives. Such input-output pairs may also be used to fine-tune ChatGPT. These may inadvertently reveal sensitive information of users. Individuals' interaction histories with ChatGPT may also be used to track and profile individuals. 
In addition, many of the databases that ChatGPT can use come from the Internet even social platforms such as Twitter, which means that ChatGPT may learn content that may leaks privacy of individuals and lacks fact-checking, and further not only generate incorrect or wrong information, but also cause privacy issues.

\subsection{Transparency}

\begin{figure}
\centerline{\includegraphics[width=18.5pc]{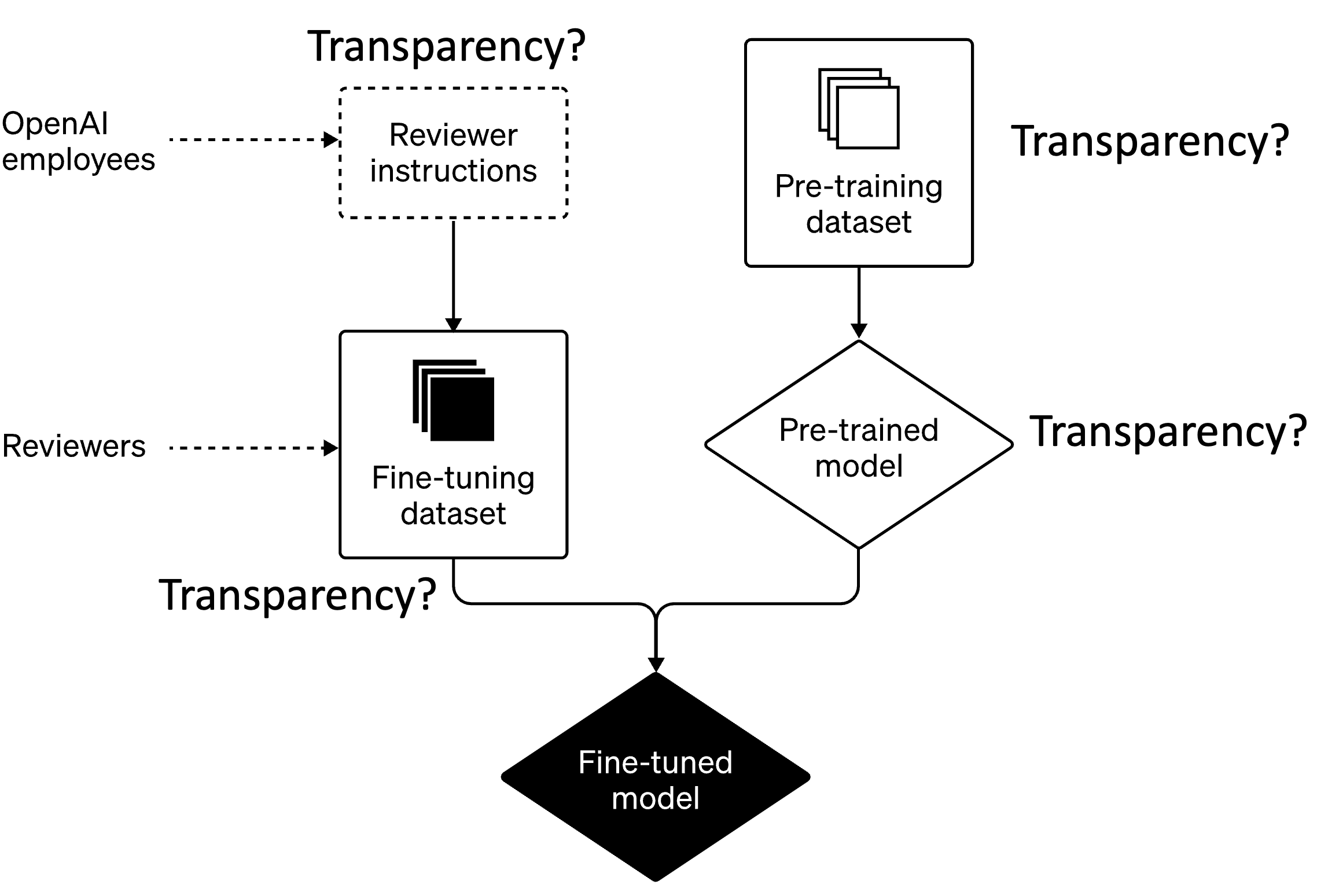}}
\caption{Two main steps in building ChatGPT (adapted from \cite{openai_2023})}\vspace*{-5pt}
\label{fig:two_steps_chatgpt}
\end{figure}

Figure~\ref{fig:two_steps_chatgpt} shows two main steps in building ChatGPT. However, OpenAI did not release much information about ChatGPT. For example, what training data is used, what are the training and testing data and their sizes, what model is used, what are the review instructions, and who are reviewers are not transparent. But OpenAI heavily emphasized its performance on question answering. 
Therefore, ChatGPT's inner workings are opaque to users, which can make it difficult to understand how it arrives at its responses. The lack of transparency affects user trust in ChatGPT the ability of users to make informed decisions about how to use the model.

Furthermore, ChatGPT's answer is ``random'' based on statistical models. The same question may give slightly different answers in differet queries with ChatGPT. The lack of awareness of randomness makes ChatGPT less trustworthy \cite{krugel_moral_2023}.

\subsection{Abuse}

The primary goal of ChatGPT is to generate seemly reasonable human-like text responses to inputs using natural language. However, trained with reinforcement learning, it currently does not have source of truth  and does not include the accuracy. 
The ability to generate human-like text could result in the misue and abuse of the technology such as spreading misinformation or impersonating individuals. 

For example, programmers acted firstly and are more likely to use such generative AI tools. StackOverflow, a popularr question-and-answer site for programmers, had already moved to ban the submission of ChatGPT-generated answers as the site explained,
``Overall, because the average rate of getting correct answers from ChatGPT is too low, the posting of answers created by ChatGPT is substantially harmful to the site and to users who are asking and looking for correct answers'' \cite{stackoverflow_2022}.
Some big financial companies also banned the use of ChatGPT in their work mainly concerning its accountability.

Furthermore, phishing email scams, online job hunting and dating scams, and even political propaganda may benefit from human-like text from ChatGPT. Just imagine, in the past, cross-border fraudulent emails were often exposed due to insufficient language translation, but with AI capable of translation and text generation, it may be even more difficult to detect.

\subsection{Authorship}

Since ChatGPT can generate human-like writings using natural language, it may be used in different situations that need text writings. For example, students may use ChatGPT in their home work or report writing. More and more academics from universities pointed out that they received text reports generated by students using ChatGPT.
They have difficulty to differentiate the authorship for the plagiarism concerns between students and AI so that teachers can evaluate students' performance objectively. 
Furthermore, ChatGPT has been used in creative work such as creative writing and music composing, which introduces not only the authorship concerns but also the human's creativity concerns.

\section{CHALLENGES}





Despite the potential of AI and ChatGPT to greatly enhance many areas of life, including communication and problem-solving in various domains, even in medicine \cite{MuellerEtAl:2021:TenCommandments}, as with any new technology, it is important to be aware of potential challenges that may arise. The use of ChatGPT in various applications has prompted the identification of the following challenges: 

\begin{itemize}
    \item[{\ieeeguilsinglright}]  {\it Blind trust} --- 
    Over-reliance on AI systems without proper validation or verification can lead to incorrect or inappropriate decision making, potentially resulting in harm to users or other negative consequences. ChatGPT lacks the source of truth of responses, the over-reliance on ChatGPT in decision making may result in unexpected harm to users. While the check of the truth of ChatGPT responses is a challenge.
    
    \item[{\ieeeguilsinglright}]  {\it Over-regulation (no guts, no glory)} --- Excessive regulation could prevent innovation and progress, as overly strict regulations could limit the ability of private and commercial users to experiment and take well-known risks with new AI technologies. ChatGPT has been demonstrating its strong capabilities since its first release. However, some counties and organisations have banned its uses in their organisations because of various adverse concerns. It is a challenge for stakeholders in the regulation of its uses. Although the regulation of the uses of ChatGPT is highly important, the over-regulation may affect the innovation progress of new technologies.
    
    \item[{\ieeeguilsinglright}]  {\it Dehumanization} --- 
    AI systems that replace human interaction and compassion in human-to-human relationships can lead to a loss of empathy and decreased satisfaction in society. The human-like responses from ChatGPT may result in the difficulty in differentiating machines and humans and thus affects human-to-human relationships. It is currently still a challenge to differentiate responses from ChatGPT and humans.
    
    \item[{\ieeeguilsinglright}]  {\it Wrong targets in optimization} --- 
    AI systems that prioritize metrics that do not align with social norms can result in social dislocations. Such norms are unwritten rules or expectations that guide behavior and interactions within a community. Social norms can be formal or informal, and they can vary based on cultural, social, and historical contexts. ChatGPT is also without exception in lacking the alignment with social norms fully to prioritize its performance metrics. It is a challenge to consider such social norms in ChatGPT.
    
    \item[{\ieeeguilsinglright}]  {\it Over-informing and false forecasting} --- 
    AI systems that generate too much information or provide false predictions can lead to confusion and decreased trust in the technology. Users can get any number of responses for one query from ChatGPT and there is no any accuracy information on those responses. Therefore, it is a challenge for ChatGPT to foster trust under such conditions.  AI systems that rely solely on statistical models without considering individual user circumstances can lead to incorrect or inappropriate actions. Responses from ChatGPT are randomly statistically generated. Different responses may be generated for a same query. ChatGPT may generate incorrect responses because of its statistical characteristics, and it is a challenge to generalise its responses. 
    
    \item[{\ieeeguilsinglright}]  {\it Self-reference (AI-based) monitoring} --- 
    AI systems that rely solely on themselves for evaluation, without independent oversight, can lead to a lack of accountability and decreased transparency in decision making. As shown in Figure~\ref{fig:two_steps_chatgpt}, ChatGPT uses both supervised and unsupervised learning to train the model. OpenAI did not open much information on how ChatGPT is evaluated and monitored despite the use of self-reference approach commonly used in the community. 
\end{itemize}

\section{RECOMMENDATIONS}

Considering significant challenges of ChatGPT as discussed above, this section provides recommendations to different stakeholders in ChatGPT for its responsible uses. We first identify various stakeholders involved in ChatGPT. Recommendations are then suggested to different stakeholders.

\subsection{Stakeholders in ChatGPT}

There are various stakeholders involved in ChatGPT. Here are some examples:

\begin{itemize}
  \item[{\ieeeguilsinglright}]  {\it Researchers and developers} --- These stakeholders are involved in developing and improving ChatGPT technologies. They may work for academic institutions, research organizations, or private companies.
  
  \item[{\ieeeguilsinglright}]  {\it Users and consumers} --- These stakeholders are the end-users of ChatGPT technologies. They may use ChatGPT for various purposes, such as information retrieval, language translation, and creative writing.
  
  \item[{\ieeeguilsinglright}]  {\it Regulators and policymakers} --- They are responsible for establishing legal and ethical guidelines for the development and use of ChatGPT technologies. They may work for government agencies, international organizations, or industry associations. These stakeholders collaborate closely with advocacy groups and civil society organizations, which represent the interests of various groups affected by ChatGPT technologies, such as privacy advocates, human rights groups, and marginalized communities. Advocacy groups may lobby for policy changes or raise public awareness about the risks and benefits of these technologies.
  
  \item[{\ieeeguilsinglright}]  {\it Ethicists and social scientists} are stakeholders who focus on the ethical and social implications of ChatGPT technologies. They may study the impact of these technologies on society, culture, and human behavior. Their role is to reflect on the developments and provide guidance on how to address any ethical or social issues that arise. While they may not directly influence the development of these technologies, their work helps ensure that ChatGPT technologies are developed and used responsibly.
\end{itemize}

\subsection{Recommendations for researchers and developers}


Considering challenges ChatGPT faces and characteristics of researchers and developers, the recommendations for researchers and developers of ChatGPT are:

\begin{itemize}
    \item[{\ieeeguilsinglright}]  {\it Do not be a algorithmic pied piper and seduce} and deliberately mislead your users. Take responsibility for providing background information about bias, privacy in an active way. If possible, offer a feature to explain why a particular statement was made in ChatGPT.
    
    \item[{\ieeeguilsinglright}]   {\it Protect the vulnerable} ---
    It is important to protect vulnerable individuals who may not fully understand the disclaimer in ChatGPT. This includes children, young people, and individuals with cognitive disabilities or lower cognitive function, who may require additional protection.
    
    \item[{\ieeeguilsinglright}]  {\it Give reasons for answers, avoid made-up sentences unless they are explicitly requested} --- This command is important for ChatGPT because it emphasises the importance of providing clear and well-justified responses to users. When the ChatGPT generates an outcome or response, it should be able to explain the reasoning behind it, rather than simply providing a result without any explanation.
    
    Providing justification for a response can help build trust and credibility with the user, and can also help the user better understand the bot's thought process and decision-making. Additionally, this command emphasises that the bot should only produce outcomes when they are deliberately requested by the user, rather than providing unsolicited responses.
    
    \item[{\ieeeguilsinglright}]  {\it Connect ChatGPT to domain knowledge} --- Connecting ChatGPT models to domain-specific knowledge representations curated by a community and/or experts can greatly enhance the accuracy and relevance of the responses provided by the model. These knowledge representations can take different forms, such as taxonomies, ontologies, or knowledge graphs, and can be both human-readable and machine-readable. By including domain-specific knowledge in the training data, the model can learn to incorporate this information into its responses. This approach may require significant domain expertise to curate and annotate the training data.
\end{itemize}

\subsection{Recommendations for users}

The recommendations for users of ChatGPT include:
\begin{itemize}
    \item[{\ieeeguilsinglright}]  {\it Check, re-check, double-check.},  if users intend to use the result of a ChatGPT conversation as fact. This is a fundamental principle of reliable science as well as trustworthy journalism that emphasises the importance of verifying information before publishing it. Before sharing any information on ChatGPT, make sure to check the source and ensure that it is credible and reliable. Avoid sharing information from unknown or unverified sources. Also be aware of own biases and those of others in the chat. Double-check any information that seems too good to be true or aligns too closely with your own beliefs.  When using ChatGPT, it is important to critically evaluate the information presented, distinguishing between fact and fiction, and considering how the responses were generated. 

    \item[{\ieeeguilsinglright}]  {\it Don't mix facts and fiction} --- To use ChatGPT responsibly, it is important to distinguish between reality and fiction and to contextualize the information obtained from the platform. While it is not necessary to rely solely on factual information, it is crucial to differentiate between statements that are part of a fictional story and those that are intended to be true. For instance, scientific statements can be presumed to be accurate, whereas statements in a work of fiction may not be. Therefore, when using ChatGPT, it is essential to put the used text into the right context. 
    
    \item[{\ieeeguilsinglright}]  {\it Don't use a result of ChatGPT that you don't understand} --- This rule emphasises the importance of understanding the meaning and implications of a statement before using it. This rule can also be applied to users of ChatGPT to help ensure that messages being sent are clear and accurately reflect the intended meaning. If you come across a technical term or jargon in a message that you are not familiar with, avoid using it in your own message. Instead, take the time to research the meaning and ensure that you understand it fully before using it.
    
    \item[{\ieeeguilsinglright}]  {\it Don't get into 'waffling' and try to convince by the sheer amount of text generated by a machine} --- emphasises the importance of being clear and concise in communication, and avoiding the use of overly complex language or convoluted sentence structures. And also do not to blind with superficiality which can be easy generated by ChatGPT. 
    
    \item[{\ieeeguilsinglright}]  {\it Do not assign ChatGPT any responsibility to you who has not explicitly accepted it in its terms and conditions} --- This rule emphasises the importance of understanding the terms and conditions of using ChatGPT. While this may be legally necessary, it also emphasises the importance of reading and understanding the terms and conditions of any platform or service before using it. 

     \item[{\ieeeguilsinglright}]  {\it Ignore emotional language} --- 
     Despite the human-like qualities, ChatGPT does not have emotions and feelings, but it can sometimes fake such emotions. Emotional language from ChatGPT could be ignored.
\end{itemize}

\subsection{Recommendations for regulators and policymakers}

The recommendations for regulators and policymakers are:

\begin{itemize}
    \item[{\ieeeguilsinglright}]  {\it Don't over regulate} --- Finding the right balance between regulation and free use can be a challenging task for regulators. On the one hand, regulation can be necessary to protect individuals and ensure fair competition. On the other hand, excessive regulation can stifle innovation and limit the benefits of new technologies or services. To strike the right balance, regulators should consider a variety of factors, including the potential risks and benefits of the technology or service, the impact of regulation on users and businesses, and the potential for self-regulation or market-based solutions. In addition, regulators should engage with stakeholders, including users, businesses, and experts, to ensure that their approach is informed by a range of perspectives. Ultimately, the goal should be to create a regulatory environment that promotes innovation and growth while protecting the public interest.

    \item[{\ieeeguilsinglright}]  {\it Thou shalt not concentrate information and communication in one place} --- Concentrating information and communication in one place can create imbalances of power and increase the potential for abuse. When one entity has exclusive control over information and communication channels, they can use that power to manipulate or exploit others. This can occur in a variety of contexts, including social media platforms, news media organizations, and government agencies but also with ChatGPT in the future. To prevent these imbalances of power, it is important to distribute information and communication across multiple platforms and systems, promoting competition and diversity. This can help to ensure that no single entity has too much control or influence over the flow of information and communication, and that individuals and groups have the freedom to express themselves and access the information they need.
\end{itemize}

\subsection{Recommendations for Ethicists}

The recommendations for ethicists are:

\begin{itemize}
    \item[{\ieeeguilsinglright}]  {\it Understand ethics roles fully in innovative technologies} --- AI ethics concerns with the human moral behaviour as they design, construct, use and treat artificially intelligent beings, as well as concerns with the moral behaviour of AI agents \cite{jobin2019global,zhou2022ai}. Ethicists need to fully understand the roles of ethics in ChatGPT in order to not only guide the ethical development of ChatGPT, but also provide guidance on the ethical use of ChatGPT more effectively.
    \item[{\ieeeguilsinglright}]  {\it Collaborate with experts closely from multiple disciplinaries} --- The conversation about ethics of ChatGPT is a philosophical discussion and needs to be elevated to a sufficiently high level from different fields. For example, legal or social experts are usually good at ethical issues related to data governance, but they may not have deep knowledge on how an AI model such as LLMs is built with a large number of parameters as AI experts be. Therefore, ethicists need to collaborate with experts that span the fields of AI, engineering, law, science, economics, ethics, philosophy, politics, and health as well as other.
\end{itemize}


\section{DISCUSSION}

LLMs such as ChatGPT can be used to fulfil typical language tasks such as summarising text paragraphs and writing news articles, answering difficult questions, generating ideas, programming computer code, writing interesting novels, as well as others \cite{floridi2020gpt}. However, it lacks a firm moral stance, and it is suggested that users do not carelessly follow ChatGPT's advice \cite{krugel2023moral}. We need to ensure individuals and organisations use the ChatGPT ethically, legally, and responsibly.

There are no widely accepted guidelines and standards for the use of ChatGPT yet. Chan \cite{chan2023gpt} argued to first focus on the regulation of professional AI developers and users by government regulatory agencies, and high-quality curated datasets for less harmful language model outputs. There is also a need to fill the gap between abstract ethical principles and  practical applications \cite{zhou2022ai}. Furthermore, public education and digital literacy are important measures to address potential intentional misuse for manipulation and unintentional harm caused by bias from language models. 

This paper proposed commandments for different stakeholders for the responsible use of ChatGPT. For example, a known limitation of ChatGPT is that it may provide answers to questions that are simply wrong. The fact-check of answers is highly important for its responsible uses. Furthermore, a feature to explain why a particular statement was made in ChatGPT will foster user trust in responses of ChatGPT. Therefore, the development of fact-check and justification of responses of ChatGPT is highly suggested for the responsible use of ChatGPT. 
Developers must also develop systems to detect and mitigate bias, ensure user privacy and security, differentiate responses from ChatGPT and humans, and prevent misuse of the technology.

In summary, ChatGPT has been demonstrating extensive applications with human-like writings. However, users have shown various ethical concerns such as intentional misuse for manipulation and unintentional harm caused by bias, because of the data it uses and opaque models as well as human-like responses.
There are still various challenges to address these ethical concerns. The commandments presented in this paper will motivate stakeholders for the ethical use of ChatGPT.

\section{CONCLUSION}

ChatGPT has been becoming a widely discussed topic because of its powerfulness in generating human-like text for various purposes since it was released in November 2022. However, there are various ethical concerns associated with the use of ChatGPT. 
This paper highlighted typical ethical concerns on ChatGPT such as Bias, Privacy, and Abuse, and articulates key challenges when ChatGPT is used in various applications. The proposed practical commandments for different stakeholders of ChatGPT can serve as guidelines for those applying ChatGPT in their applications. The future of this work will focus on the development of guidelines and tools for fact-check and justification of responses for the responsible use of ChatGPT.


\section{ACKNOWLEDGMENTS}
The Acknowledgments is always plural even if there is a single acknowledgment. The author(s) would like to thank A, B, and C. This work was supported by XYZ under Grant \#\#\#.

\def\refname{REFERENCES}

\bibliographystyle{IEEEtran}
\bibliography{IEEEcsMag}

\end{document}